\documentclass[10pt,twocolumn,letterpaper]{article}

\usepackage{cvpr}
\usepackage{times}
\usepackage{epsfig}
\usepackage{graphicx}
\usepackage{amsmath}
\usepackage{amssymb}
\usepackage[utf8]{inputenc}
 \usepackage{mathrsfs}
\usepackage{graphicx}
\usepackage{subfig}
\usepackage{xcolor}
\usepackage{booktabs}
\usepackage{pbox}
\usepackage{array}
\usepackage{makecell}
\usepackage{enumitem}

\usepackage[font=small,skip=2pt]{caption}

\newcommand{\stitle}[1]{\vspace{5pt} \noindent\textbf{#1.}\ }

\newtheorem{proposition}{Proposition}
\newtheorem{remark}{Remark}

\newcommand{\esm}[1]{\ensuremath{#1}}

\newcommand{\ms}[1]{\esm{\mathsf{#1}}}

\newcommand\reals{\ms{R}}

\newcommand\sparam{\alpha}

\newcommand\pathfn{\gamma}

\newcommand\synteq{::=}

\newcommand\integratedgrads{\ms{IG}}
\newcommand\blurintegratedgrads{\ms{BlurIG}}

\newcommand\pathintegratedgrads{\ms{PathIntGrads}}

\newcommand\methodname{\ms{BlurIG}}

\usepackage[pagebackref=true,breaklinks=true,letterpaper=true,colorlinks,bookmarks=false]{hyperref}

\cvprfinalcopy % *** Uncomment this line for the final submission

\begin{document}

%%%%%%%%% TITLE
\title{Attribution in Scale and Space}

%TODO(Mukund) Can you please verify institutional affiliation. 
\author{
% We can also do this.
%\vspace{-0.5cm}
%Shawn Xu \hspace{0.5cm} Subhashini Venugopalan \hspace{0.5cm} Mukund Sundararajan\\
%Google LLC\\
%{\tt\small \{jinhuaxu,vsubhashini,mukunds\}@google.com}
Shawn Xu\\
Google AI Healthcare\\
{\tt\small jinhuaxu@google.com}
\and
Subhashini Venugopalan\\
Google Research\\
{\tt\small vsubhashini@google.com}
%Google Inc.\\
\and
Mukund Sundararajan\\
Google Inc.\\
{\tt\small mukunds@google.com}
}

\maketitle
\begin{abstract}
We study the attribution problem~\cite{SVZ13} for deep networks applied to perception tasks. For vision tasks, attribution techniques attribute the prediction of a network to the pixels of the input image. 
We propose a new technique called \emph{Blur Integrated Gradients (\methodname{})}. This technique has several advantages over other methods. First, it can tell at what scale a network recognizes an object. It produces scores in the scale/frequency dimension, that we find captures interesting phenomena. Second, it satisfies the scale-space axioms~\cite{Lindeberg}, which imply that it employs perturbations that are free of artifact. We therefore produce explanations that are cleaner and consistent with the operation of deep networks. Third, it eliminates the need for a `baseline' parameter for Integrated Gradients~\cite{STY17} for perception tasks. This is desirable because the choice of baseline has a significant effect on the explanations. 
We compare the proposed technique against previous techniques
%ly proposed techniques 
and demonstrate application on three tasks: ImageNet object recognition, Diabetic Retinopathy prediction, and AudioSet audio event identification. Code and examples are on github\footnote{{\tt \href{https://github.com/PAIR-code/saliency}{https://github.com/PAIR-code/saliency}}}.
\end{abstract}

%----------------------------------

\section{Introduction}
\label{sec:intro}
 There is considerable literature on feature-importance/attribution for deep networks~\cite{BSHKHM10,SVZ13,SGSK16,BMBMS16,SDBR14,LL17, STY17,Lundberg2017AUA}. An attribution technique distributes the prediction score (\eg sentiment or prevalence of disease) of a model for a specific input (\eg paragraph of text or image) to its base features (\eg words or pixels); the attribution to a base feature can be interpreted as its contribution to the prediction. 
 
 Suppose we have a model that predicts diabetic retinopathy (DR) from an image of the \emph{fundus} of the eye. The attributions identify the (signed) importances of each pixel to the prediction. This tells us what part of the eye (\eg retina/optic disc/macula), the network considers important to the prediction. We can use these attributions to debug the network---is the prediction based on a known pathology of diabetic retinopathy. The attributions could also assist the doctor. If the network's prediction differs from the doctor's the attributions could help explain why. This could improve diagnosis accuracy. \cite{jama-dr} elaborates the application attributions to DR.     
 
 In this work, we study the attribution problem for perception tasks, i.e., the input is an image or a waveform and the features are pixels or time points. Perception tasks are different from others tasks (natural language, drug discovery, recommender systems): The base features (pixels or time points) are never influential by themselves; information is almost always contained in higher-level features like edges, textures or frequencies. This makes perception tasks worthy of separate study.
    
% \subsection{Our Contributions}
% \label{sec:contrib}
 
\textbf{Our Contributions.} We use the theory of scale-space \cite{scale-space} to make two contributions to the attribution literature:
 
 %\begin{itemize}
\textbf{Explanation in scale and space.} Previous attribution techniques produced feature importance for pixels, \ie, points in space. They do not produce localization in frequency. We propose a technique called Blur Integrated Gradients (\methodname{}) to produce explanations in both scale/frequency and space. We can therefore tell that the detection of a steel-arch bridge is based on coarse, large-scale features, whereas the detection of a dog-breed depends on fine-grained features (see Figure~\ref{fig:steel_arch_bridge_prediction_trend}). (This is not simply a statement that dogs are smaller than bridges. In the dataset we study, the images frame dogs and bridges similarly.)

%\textbf{Perturbations, artifacts, and scale-space axioms.}
\textbf{Perturbations free of artifacts.} All attribution techniques involve perturbations of either the inputs (e.g.~\cite{STY17}) or the network state (e.g.~\cite{SVZ13}). The premise is that \emph{removing} an important feature causes a large change in the prediction score. The literature does not however discuss if these perturbations could accidentally \emph{add} features. Then, the change in the score could be because a different object is detected, and not because information is destroyed. This could result in the explanation artifacts that identify influential features that are not actually present in the input. (Compare the IG and Blur IG explanations for `starfish' in Figure~\ref{fig:comparison-of-methods}.) We discuss how the scale space-theory addresses accidental feature creation. 
 %\end{itemize}

%----------------------------------

\section{Related Work}
 \label{sec:related}
 There is considerable literature on feature-importance/attribution for deep networks, i.e. techniques to identify the influence/importance of each feature to the prediction of a network for a specific input. Some techniques perform this attribution by propagating the prediction from the output back towards the input (\eg \cite{SVZ13, Lundberg2017AUA, SGSK16,GuidedBackProp}). Other techniques build upon the gradient of the prediction with respect to the input or a coarse version of the input (\eg \cite{Selvaraju,STY17,XRAI}). Our work implicitly addresses a critique of Integrated Gradients identified in~\cite{XRAI}---that the explanations depend on the auxiliary baseline parameter, and varying the baseline changes the explanation. ~\cite{XRAI} addresses this criticism by averaging the attributions over two baselines (a black and a white image). In contrast the technique we propose subsumes the baseline; there is no need for this parameter. 
 
 There are also other techniques like~\cite{Lime} that are not specific to deep networks. Another class of techniques uses gradient ascent to modify the input to emphasize features critical to a neuron activation, either for an output neuron, or an internal one~\cite{deepdream}. \cite{fong} formulates an optimization problem to highlight regions of an image that are important to a certain prediction. They incorporate a smoothness criterion based on Gaussian blur within the optimization objective to produce smooth, contiguous masks. Though the explanations have higher verisimilitude for real-world images, it is possible that they do not pick up fine features or textures, aspects that are critical to medical diagnosis applications (\eg Section~\ref{sec:DR}).  
 
 None of the works above address attribution along the scale/frequency dimension, or ensure that the perturbations don't manifest new information. 
 We borrow concepts from the literature on scale-space theory~\cite{Koenderink, Lindeberg, Witkin}; this is a theory of multi-scale signal representation that represents the an image or signal as a one-parameter family of smoothed images or signals. We use the prediction gradients of this family of images to produce explanations. The smoothing process is axiomatically guaranteed \emph{not} to produce artifacts in the image(see Section~\ref{sec:causality}) using the so called scale-space axioms~\cite{Lindeberg}. Therefore the resulting explanations will also be free of artifacts.   There is a large literature that uses scale-space theory and Gaussian blurs to detect blobs, edges or other higher-level features \eg~\cite{SIFT, Blob}; this literature is unrelated to deep networks or the interpretability/explanation/feature importance problem.
 
 Finally, we mention a brief connection to cooperative game theory. Integrated Gradients~\cite{STY17} that we build on, is itself based on a method called Aumann-Shapley~\cite{AS74}. The method is constructed axiomatically using axioms that characterize desirable properties of attribution techniques \eg~\cite{Friedman}. These axioms differ from the scale-space axioms mentioned earlier. The scale-space axioms axiomatize the type of perturbation.

%----------------------------------

 \section{Attribution in Scale and Space}
 \label{sec:scale-and-space}
 
 \subsection{Blur Integrated Gradients (\methodname{})} 
 \label{sec:Blur IG}
 
\methodname{} extends the Integrated Gradients~\cite{STY17} technique. %We extend a technique called Integrated Gradients~\cite{STY17}.
Formally, suppose we have a
function $F: \reals^{m \times n} \rightarrow [0,1]$ that represents a deep network.
Specifically, let $z(x,y) \in \reals^{m \times n}$ be the 2D input at hand, and
$z'(x,y) \in \reals^{m \times n}$ be the 2D baseline input, meant to represent an informationless input. For vision networks, this could be the black image, or an image consisting of noise. We consider the straight-line path (in $\reals^{m \times n}$) from the baseline $z'$ to the input
$z$, and compute the gradients at all points along the path. The path can be parameterized as
\begin{equation*}
\gamma(x, y, \alpha) = z'(x,y) + \alpha \cdot (z(x,y) - z'(x,y))
\end{equation*}
Integrated gradients (IG) are obtained by accumulating these gradients. The integrated gradient for an input $z$ and baseline
$z'$ is:
\begin{equation}\label{eq:ig}
\integratedgrads(x, y) \synteq (z(x,y)-z'(x,y)) \cdot \int_{\sparam=0}^{1} \tfrac{\partial F(\gamma(x, y, \alpha))}{\partial \gamma(x, y, \alpha)  }~d\alpha
\end{equation}

Let us call this the \textit{intensity scaling} IG; if you use a black image as the baseline $z'$, the path scales the intensity of the image. Let us instead consider the path defined by successively blurring the input by the Gaussian blur filter. Formally, let
\begin{equation*}
    L(x,y,\sparam) = \sum_{m=-\infty}^{\infty} \sum_{n=-\infty}^{\infty} \frac{1}{\pi \sparam} e^{-{\frac{x^2+y^2}{\sparam}}} z(x - m, y -n)
\end{equation*}
be the discrete convolution of the input signal with the 2D Gaussian kernel with variance $\sparam$, also known as the scale parameter. Blur integrated gradients is obtained by accumulating the gradients along the path defined by varying the $\sparam$ parameter:
\begin{equation}\label{eq:blurig}
\blurintegratedgrads(x, y)\synteq \int_{\sparam=\infty}^{0} \tfrac{\partial F(L(x,y,\sparam))}{\partial L(x,y,\sparam)} \tfrac{\partial L(x,y,\sparam)}{\partial \sparam}~d\sparam
\end{equation}

Implementation-wise, the integral can be efficiently approximated using a Riemann sum:
\begin{equation*}
    \blurintegratedgrads(x, y) \approx \sum_{i=1}^{s} \tfrac{\partial F(L(x,y,\sparam_i))}{\partial L(x,y,\sparam_i)} \tfrac{\partial L(x,y,\sparam_i)}{\partial \sparam_i} \tfrac{\sparam_\text{max}}{s}
\end{equation*}
where $\sparam_i = i \cdot \tfrac{\sparam_{\text{max}}}{s}$ and $s$ is the number of steps in the Riemann approximation. The gradients are obtained using numerical differentiation. The maximum scale $\sparam_\text{max}$ should be chosen to be large enough so that the resulting maximally blurred image is information-less. Small features are destroyed at smaller scales. But the specific value of $\sparam$ at which a feature is destroyed depends on the variance of the intensity of the pixels that form the feature (see Figure 5.10 from~\cite{Lindeberg} for a discussion of this). The smaller the variation, the smaller the $\alpha$ at which it is destroyed.

Note that we have defined blur integrated gradients for 2D signals, but it exactly equivalent (after reparameterization) for the 1D case.

 \subsection{Interpretation of Blur IG}
 \label{sec:interpretation}
 It is well known that $L(x,y, \sparam)$ in equation~\ref{eq:blurig} is the solution of the 2D diffusion equation:
 \begin{equation*}
    \tfrac{\partial L}{\partial \sparam} = \tfrac{1}{4} \nabla^2 L
 \end{equation*}
 with the initial condition $L(x, y, 0) = z(x,y)$. By commutativity of differentiation with convolution,
 \begin{equation*}
    \nabla^2 L = \nabla^2 (G * z) = (\nabla^2 G) * z
 \end{equation*}
 where $G(x,y,\sparam)$ is the Gaussian kernel and $\nabla^2 G(x,y,\sparam)$ is the Laplacian of Gaussian (LoG) kernel. Thus, equation 3 can be alternatively expressed as:
 \begin{equation}
\blurintegratedgrads \synteq \tfrac{1}{4} \int_{\sparam=\infty}^{0} \tfrac{\partial F(L)}{\partial L} \cdot (\nabla^2 G) * z ~d\sparam
\end{equation}
 The LoG kernel is a band-pass filter used to detect edges at the specified scale. Thus, the first term of the integrand, which consists of model gradients with respect to the blurred image at the appropriate scale provides localization in space. This when filtered by the LoG-filtered image provides localization in frequency.
 
%\begin{figure}
%      \centering
% \begin{center}
% \begin{tabular}{cl } 
%  {\centering$\sparam$} & \hspace{0.8cm} 6.45 \hspace{1.2cm} 3.3  \hspace{1.6cm} 0.3 \\
%  {\centering$L$} & \includegraphics[width=0.375\textwidth]{figures/jackfruit_blur_path.jpeg} \\ 
%  {\centering$(\nabla^2 G) * z$} & \includegraphics[width=0.375\textwidth]{figures/jackfruit_gauss_grad.jpeg} \\ 
%  {\centering partial Blur IG} & \includegraphics[width=0.375\textwidth]{figures/jackfruit_partial_big.jpeg} \\ 
% \end{tabular}
% \end{center}
%      \caption{Aspects of Blur IG at three levels of Blur ($\alpha$). First row is the input. Second row is the Laplacian. Third row is the Blur IG integration (summation) up to that $\alpha$. }
%      \label{fig:jackfruit}
%\end{figure}

\begin{figure}%
    \centering
    \subfloat[$ \sparam=6.45 \hspace{0.7cm} \sparam=3.3  \hspace{1.3cm} \sparam=0.3 $ ]{{\includegraphics[width=0.375\textwidth]{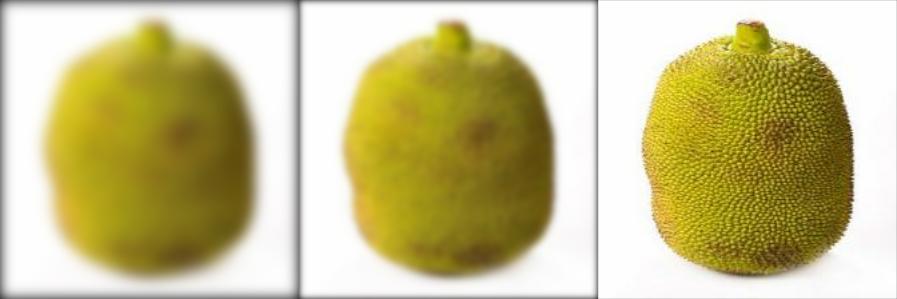}}}\\
    \vspace{-0.2cm}
    \subfloat[$\alpha = (\nabla^2 G) * z$]{{\includegraphics[width=0.375\textwidth]{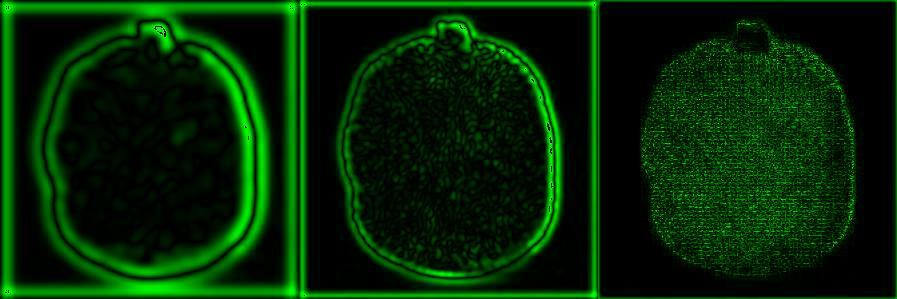} }} \\  \vspace{-0.2cm}
    \subfloat[partial Blur IG]{{ \includegraphics[width=0.375\textwidth]{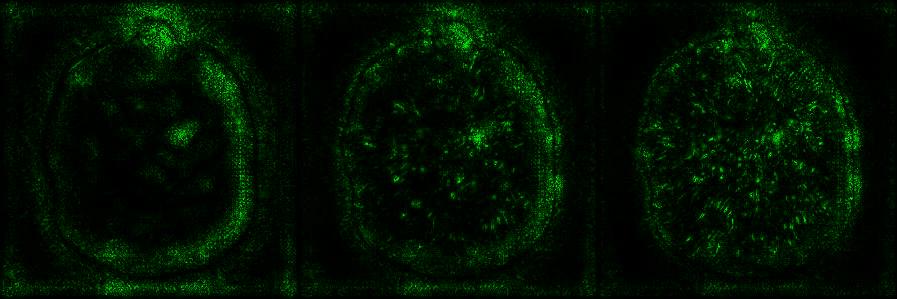}}}%
    \caption{Aspects of Blur IG at three levels of Blur ($\alpha$). First row is the input. Second row is the Laplacian. Third row is the Blur IG integration (summation) up to that $\alpha$. %Green indicates positive attributions and red the negative attributions.
    }%
    \label{fig:jackfruit}%
\end{figure}

%\todo{Shawn: Make the table above prettier, print prediction scores, and use clearer visualization.}

Figure~\ref{fig:jackfruit} shows the different components of the integrand at different scales along the blur path while explaining the `jackfruit' prediction for the Inception-Resnet-v2 model trained on ImageNet. At low resolution, the model picks up high-level details of the jackfruit, such as the stem and overall shape, while at high resolutions, the model picks up the spikiness of the texture.

%----------------------------------

 \section{Applying Scale-Space Axioms to Attribution}
  \label{sec:axioms}
 
 Attributions and Explanations are based on perturbations. Every method---for instance, Gradcam, IG, or Blur IG ---prescribes a specific set of perturbations to the input (\eg gradient computation). If the perturbations destroy `information', then the resultant change in prediction can be interpreted as feature importance; this is the desired interpretation. However, if the perturbation creates information, then the resultant change in score is not due to a feature present in the input, and the result will be a misleading, uninterpretable explanation.

 In this section, we use the scale-space theory~\cite{Witkin, Koenderink} to identify sequences of input perturbations that provably do not manifest information. 
 
 \subsection{The Causality Axiom}
 \label{sec:causality}

 Consider a sequence of grayscale images $L:R^3 \rightarrow R$; here $L(x,y,t)$ denotes the intensity of pixel $x,y$ at scale $t$. The input image is $L(x,y,0)$. As the scale parameter $t$ increases, so does the level of perturbation. Scale-space theory formalizes `Non-Creation of New Structure'\cite{Lindeberg} using the following `Causality' axiom\cite{Koenderink}: 
 
 \textbf{Causality:} Local extrema should not be enhanced as you increase the scale parameter. Namely, if $L(x,y,t)$ is at a local maxima (resp. minima) in $x,y$, then $L$ should not increase (resp. decrease) with $t$. Equivalently, no new level surfaces are created as the scale parameter increases.
 Visual features like edges or textures correspond to local extrema in the representation $L(x,y,\cdot)$. Therefore, the causality axiom states that features should only be destroyed as $t$ increases. In fact, the axiom states something stronger: that existing extrema are not enhanced by the path.

% \begin{figure}
%     \begin{minipage}[t]{0.33\columnwidth}
%       \includegraphics[width=\linewidth]{figures/x2_blur.png}
%         \caption{\\ Gaussian Blur \\ scale-space for $x^2$}
%         \label{fig:blur}
%     \end{minipage}\hfill
%     \begin{minipage}[t]{0.33\columnwidth}
%   \includegraphics[width=\linewidth]{figures/x2_intensity_zero.png}
%         \caption{\\ Intensity scaling \\ (zero baseline \\ aka black baseline) \\ scale-space for $x^2$.}
%     \label{fig:intensity}
%     \end{minipage}
%     \begin{minipage}[t]{0.33\columnwidth}
%   \includegraphics[width=\linewidth]{figures/x2_intensity_random.png}
%         \caption{\\ Intensity scaling \\ (random baseline) \\ scale-space for $x^2$.}
%     \label{fig:intensity_random}
%     \end{minipage}
% \end{figure}
 
\begin{figure}
    \centering
    \subfloat[Gaussian blur]{\includegraphics[width=0.32\columnwidth]{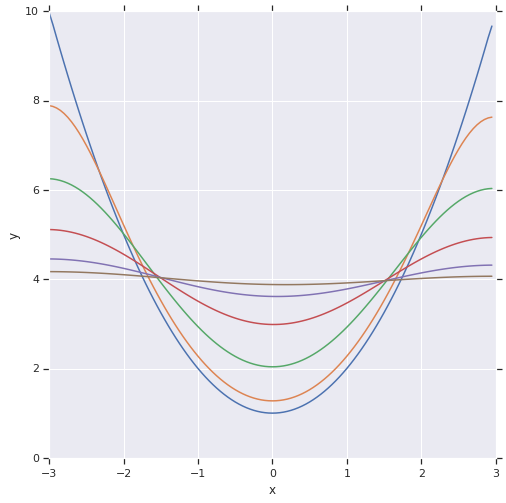}\label{fig:blur}}
    \subfloat[][Intensity scaling \ \\ (black baseline)]{\includegraphics[width=0.32\columnwidth]{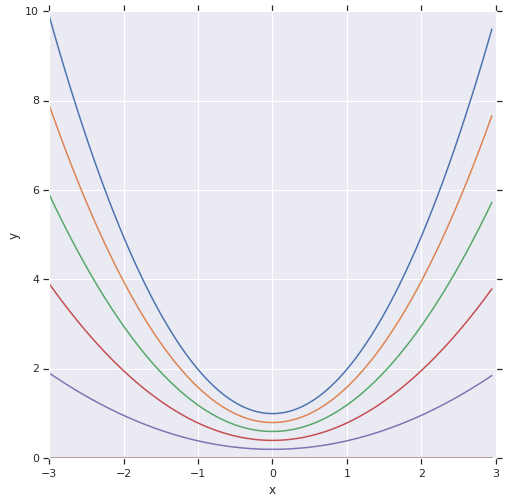}\label{fig:intensity}}
    \subfloat[][Intensity scaling\\  (random baseline)]{\includegraphics[width=0.32\columnwidth]{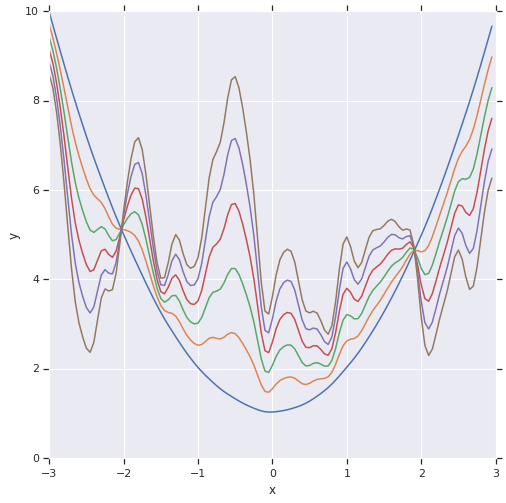}\label{fig:intensity_random}}
    \caption{Scale space for $x^2+1$ along the Gaussian blur, and intensity scaling (black and random baseline) paths.}
    \label{fig:x2_scale_space}
\end{figure}

Figure~\ref{fig:x2_scale_space} shows %s~\ref{fig:blur}, \ref{fig:intensity}, and \ref{fig:intensity_random} show 
the scale space for blur scaling and intensity scaling (both black baseline and random baseline) for the univariate function $x^2 + 1$. Blur satisfies causality---no new extrema are created and the only minima is diminished. Intensity scaling (black baseline) breaks the stronger form of causality---the only minima is enhanced. Intensity scaling (random baseline) breaks the weaker form of causality as well -- not only can the minima be diminished, but new extrema are introduced.

 \begin{remark}[Baseline choice in IG]
 \label{rem:baseline}
 If we use a black image as a baseline (as suggested by~\cite{STY17}) for IG, we get a series of images that differ in intensity. As Fig.~\ref{fig:intensity} shows, minima can be enhanced along this path. However it is easy to see that no new extrema are created by intensity scaling. In this sense, it satisfies a weaker notion of causality axiom. However, as discussed by~\cite{XRAI}, using the black image as baseline suppresses intuitively important features that have low intensity. One remedy is to use a noise image as a baseline, as is done in~\cite{AnkurGit}. However this can create new extrema (Fig.~\ref{fig:intensity_random}), that result in spurious features (the `starfish' example from Fig.~\ref{fig:comparison-of-methods} is an instance of this.) This is somewhat mitigated by averaging the attributions over several runs of the method. But this is a heuristic fix, and involves a computational hit. Blur IG appears to be a more systematic fix.
 \end{remark}

\subsection{Justifying Blur Integrated Gradients}
 \label{sec:justify}  
 
\subsubsection{Path Methods}
\label{sec:path_methods}

IG and Blur IG aggregate the gradients along a series of images that culminate at the  
input. Clearly, there are many other possible paths, and each such path will yield a different attribution method. We first justify why path methods are desirable. 

Formally, let $\pathfn = (\pathfn_1, \ldots, \pathfn_n): [0,1]
\rightarrow \reals^n$ be a smooth function specifying a path in
$\reals^n$ from the baseline $z'$ to the input $z$, i.e.,
$\pathfn(0) = z'$ and $\pathfn(1) = z$.
Given a path function $\pathfn$, \emph{path integrated gradients} are obtained by
integrating the gradients along the path $\pathfn(\sparam)$
for $\sparam \in [0,1]$. Formally, path
integrated gradients for an input $z$
is defined as follows.
\begin{equation}
\pathintegratedgrads^{\pathfn}(z,z') \synteq \int_{\sparam=0}^{1} \tfrac{\partial F(\pathfn(\sparam))}{\partial \pathfn(\sparam)  }~\tfrac{\partial \pathfn(\sparam)}{\partial \sparam}  ~d\sparam
\end{equation}
Attribution methods based on path integrated gradients are collectively
known as \emph{path methods}. 
Notice that integrated gradients is a path method
for the straightline path specified by $\pathfn(\sparam) = z' + \sparam\cdot(z-z')$
for $\sparam \in [0,1]$.
More interestingly, path methods are the only methods
that satisfy certain desirable axioms. (For formal definitions of the
axioms and proof of Proposition~\ref{prop:path}, see
Friedman~\cite{Friedman}.)

\stitle{Axiom: Dummy} If the
function implemented by the deep network does not depend (mathematically)
on some variable, then the attribution to that variable is always
zero.

\stitle{Axiom: Linearity}
Suppose that we linearly composed two deep networks modeled by the
functions $f_1$ and $f_2$ to form a third network that models the
function $a\cdot f_1 + b\cdot f_2$, \ie, a linear
combination of the two networks. Then we'd like the attributions for
$a\cdot f_1 + b\cdot f_2$ to be the weighted sum of the attributions
for $f_1$ and $f_2$ with weights $a$ and $b$
respectively. Intuitively, we would like the attributions to preserve
any linearity within the network.

\textbf{Completeness} if for every explicand $z$, and baseline $z'$, the attributions add up to the difference $f(z) - f(z')$, the difference in prediction score for the input and the baseline.

\textbf{Affine Scale Invariance (ASI)} if the attributions are invariant under a simultaneous affine transformation of the function and the features. That is, for any $c, d$, if $f_1(z_1, \ldots, z_n) =
f_2(z_1, . . . ,(z_j - d)/c, . . . , z_n)$, then for all $j$ we have $\text{attr}_j(z, z', f_1) = \text{attr}_j((z_1, \ldots , c*z_j + d, \dots z_n),(z'_1, \ldots , c*z'_j + d, \dots z'_n), f_2)$, where $z_j$ is the value of the $j$-th pixel. ASI conveys the idea that the zero point and the units of a feature should not determine its attribution; this is defensible for machine learning  where the signal is usually carried by the covariance structure of the features and the response variable. 

\begin{proposition}(Theorem~1~\cite{Friedman})
    \label{prop:path}
  Path methods are the only
  attribution methods that always satisfy 
  Dummy, Linearity, Affine Scale Invariance and Completeness.
\end{proposition}

We now justify why the path used by Blur IG is superior to other paths for perception tasks using scale-space theory. Let $z(x,y)$ be the signal and $k(x, y, t)$ for all $t > 0$ be the family of infinitely differentiable, rapidly decreasing kernels such that the discrete convolution:
\begin{equation*}
    L(x,y,t) = \sum_{m=-\infty}^{\infty} \sum_{n=-\infty}^{\infty} k(m,n,t) z(x - m, y - n)
\end{equation*}

represents the path for $t>0$, with the initial condition $L(x,y,0) = z(x,y)$. Kernels are linear, shift-invariant. A kernel is symmetric if $k(x, y,t) = k(-x, y,t)$ and $k(x, y,t) = k(y, x,t)$. Symmetry ensures that the transformation is identical in every direction in the $x,y$ plane.
A kernel satisfies the semigroup property if $k(\cdot, \cdot,t_1) * k(\cdot, \cdot,t_2) = k(\cdot, \cdot, t_1 + t_2)$. The semigroup property ensures that all scales are treated identically. We then have the following proposition
\begin{proposition}(Theorem~3.2~\cite{Lindeberg})
    \label{prop:gaussian}
The only kernel method that is symmetric, satisfies the semi-group property, satisfies continuity in the scale parameter, and causality, is the Gaussian kernel. 
\end{proposition}

\begin{remark}
We briefly contrast the axiomatization of IG from~\cite{STY17} to this axiomatization of Blur IG. The axiomatization for IG built on top of Proposition~\ref{prop:path}. It used an additional axiom of Symmetry, that variables symmetric in the function, with equal value in the input and the baseline get identical attribution. This is a condition on the \emph{function} implemented by the deep network. Notice that the condition says nothing about the location (in $x,y$) of the two features. The condition only appears to make sense where there is no `geometry' to the input, for instance when the task is a natural language one. 
In contrast, Blur IG uses Proposition~\ref{prop:gaussian} (in addition to Proposition~\ref{prop:path}), which uses axioms about the transformations that generate the path of images. These conditions on the transformation crucially rely on the geometry of the input. for instance the condition that makes a kernel symmetric, or what makes a point a extrema in the definition of causality. 
\end{remark}
%----------------------------------

\section{Applications}
 \label{sec:applications}
 
 \subsection{Object Classification}
 \label{sec:object}
\begin{figure}
 \captionsetup[subfigure]{labelformat=empty}
 \subfloat{
   {$\qquad$ Original \hspace{0.5cm} GradCAM \hspace{0.9cm} IG \hspace{0.9cm} Blur IG}
%   %}   %%% 0.75in is half of figure height 1.5in
}\\\vspace{-0.6cm}\\
\subfloat{
   %\raisebox{-1in}{
   \rotatebox[origin=t]{90}{$\qquad\qquad$windsor}
%   %}   %%% 0.75in is half of figure height 1.5in
}
 \subfloat{
   \includegraphics[width=0.24\columnwidth]{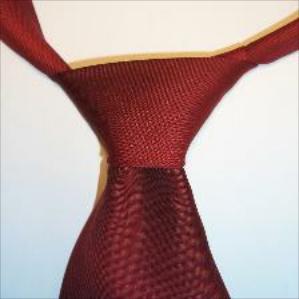}
}\hspace{-1em}
 \subfloat{
   \includegraphics[width=0.24\columnwidth]{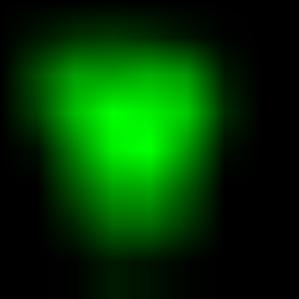}
}\hspace{-1em}
 \subfloat{
   \includegraphics[width=0.24\columnwidth]{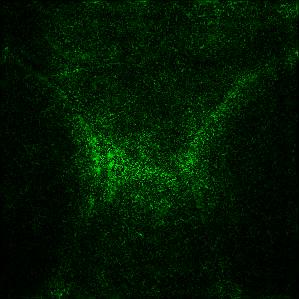}
}\hspace{-1em}
 \subfloat{
   \includegraphics[width=0.24\columnwidth]{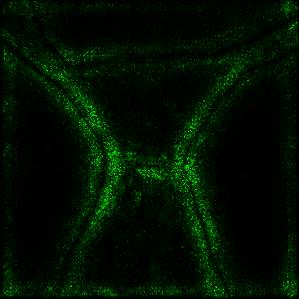}
}
\\\vspace{-1.42cm}\\
\subfloat{
   %\raisebox{-1in}{
   \rotatebox[origin=t]{90}{$\qquad\qquad$starfish}
%   %}   %%% 0.75in is half of figure height 1.5in
}
 \subfloat{
   \includegraphics[width=0.24\columnwidth]{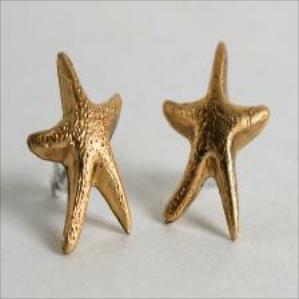}
}\hspace{-1em}
 \subfloat{
   \includegraphics[width=0.24\columnwidth]{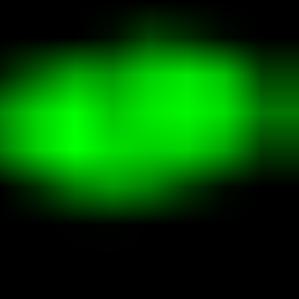}
}\hspace{-1em}
 \subfloat{
   \includegraphics[width=0.24\columnwidth]{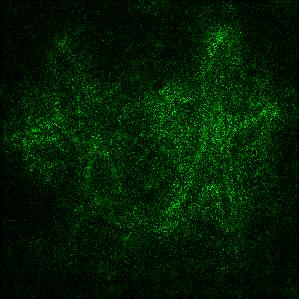}
}\hspace{-1em}
 \subfloat{
   \includegraphics[width=0.24\columnwidth]{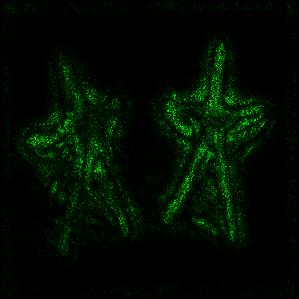}
}
\\\vspace{-1.35cm}\\
\subfloat{
   %\raisebox{-1in}{
   \rotatebox[origin=t]{90}{$\qquad\qquad$fur coat}
%   %}   %%% 0.75in is half of figure height 1.5in
}
 \subfloat{
   \includegraphics[width=0.24\columnwidth]{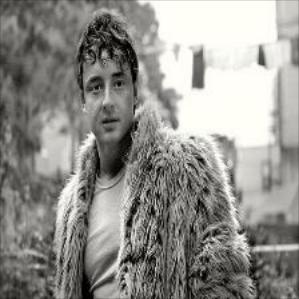}
}\hspace{-1em}
 \subfloat{
   \includegraphics[width=0.24\columnwidth]{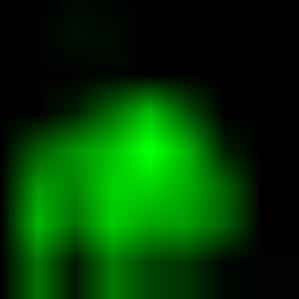}
}\hspace{-1em}
 \subfloat{
   \includegraphics[width=0.24\columnwidth]{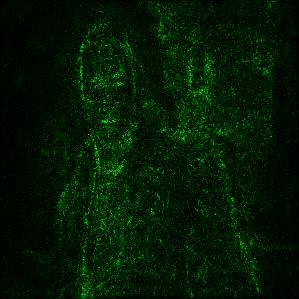}
}\hspace{-1em}
 \subfloat{
   \includegraphics[width=0.24\columnwidth]{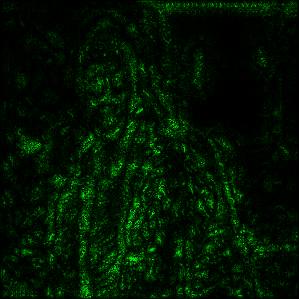}
}
\\\vspace{-1.4cm}\\
\subfloat{
   %\raisebox{-1in}{
   \rotatebox[origin=t]{90}{$\qquad\qquad$eggnog}
%   %}   %%% 0.75in is half of figure height 1.5in
}
 \subfloat{
   \includegraphics[width=0.24\columnwidth]{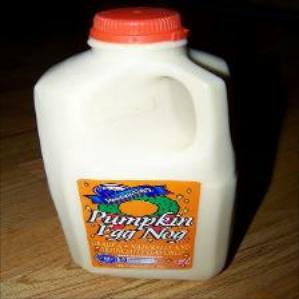}
}\hspace{-1em}
 \subfloat{
   \includegraphics[width=0.24\columnwidth]{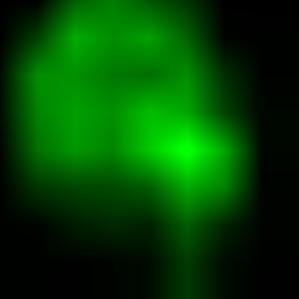}%{figures/eggnog_gradcam.jpeg}
}\hspace{-0.2cm}
 \subfloat{
   \includegraphics[width=0.24\columnwidth]{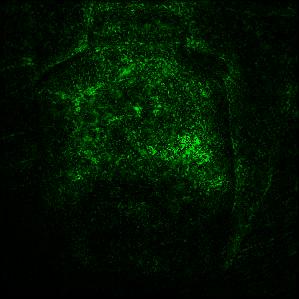}
}\hspace{-1em}
 \subfloat{
   \includegraphics[width=0.24\columnwidth]{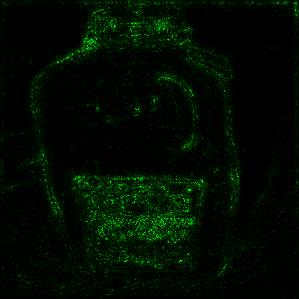}
}
\\\vspace{-1.1cm}
\caption{Comparison of GradCAM, random baseline IG, and blur IG on a sample of images from ImageNet . %Green and red colors indicate positive and negative attributions respectively.
}
\label{fig:comparison-of-methods}
\end{figure}

 %We apply Blur IG to the ImageNet object classification task~\cite{ILSVRC15}. Specifically, we study a  model that uses the Inception-Resnet-V2 architecture~\cite{Inception-v2} trained on ImageNet data. The data has natural scene images across 1000 labels. 
We apply Blur IG to %Inception-v1, and 
the Inception-Resnet-V2 architecture~\cite{Inception-v2} trained on the ImageNet classification challenge~\cite{ILSVRC15}. Figure \ref{fig:comparison-of-methods} shows the comparison of three different explanation techniques -- GradCAM~\cite{Selvaraju}, IG~\cite{STY17} (random baseline), and Blur IG -- applied to various example images. Notice that GradCAM sometimes produces an attribution mask too coarse in resolution to capture the morphology of the object (\eg windsor, starfish in Fig.~\ref{fig:comparison-of-methods}), and random baseline IG sometimes results in spurious artifacts outside of relevant regions (\eg starfish, fur coat in Fig.~\ref{fig:comparison-of-methods}); %\ref{fig2:starfish_intensity_random}, \ref{fig2:fur_coat_intensity_random});
Remark \ref{rem:baseline} discusses this issue.
%We now discuss a case where Blur IG seems strictly superior to the other methods. See 
In the last row of Figure \ref{fig:comparison-of-methods} for a container correctly predicted as `eggnog', only Blur IG attributes the packaging label correctly, while GradCAM and IG attribute the top portion of the container. It is of course possible that the top portion is sufficient to discriminate `eggnog'. However, we found that the model classifies images of containers of similar shape and color, but without the packaging label, as `water jug'. Therefore it appears that the critical feature, the packaging label, is missed by the techniques, and captured by Blur IG.
 
 %\todo{Shawn: do you want to deal with black baseline?}
 
\begin{figure}
  \subfloat[`maltese dog' trend]{
        \includegraphics[width=0.45\columnwidth]{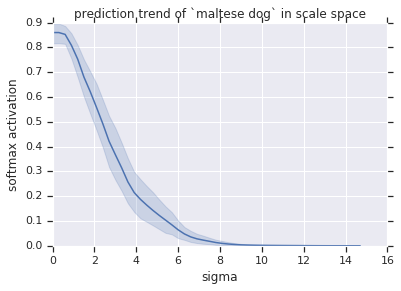}
        \label{fig:maltese_dog_prediction_trend}}
  \subfloat[`steel~arch~bridge' trend]{
        \includegraphics[width=0.45\columnwidth]{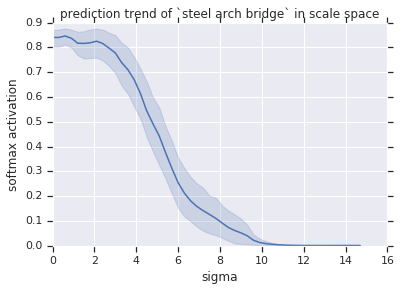}
        \label{fig:steel_arch_bridge_prediction_trend}}
    \captionsetup{justification=justified}
    \caption{Comparison of the prediction trends of `maltese dog' vs `steel arch bridge' along the blur integration path. The bulk of the prediction weight for `maltese dog' accumulates between $\sigma=0$ and $\sigma=3$, whereas the bulk of the prediction weight for `steel arch bridge' accumulates between $\sigma=3$ and $\sigma=6$. This matches our hypothesis that the model can recognize the bridge at course scale/low frequency, whereas it requires finer scale / high frequency details to recognize the dog breed (from other breeds). }
    \label{fig:prediction_trends}
\end{figure}

\begin{figure}
  \subfloat[maltese dog]{
   \includegraphics[width=0.24\columnwidth]{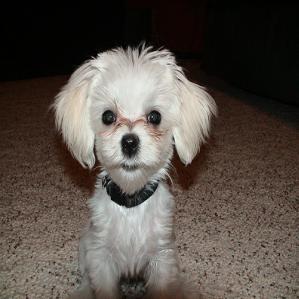}
   \label{fig:maltese_dog}
   }
  \subfloat[Blur IG]{
  \includegraphics[width=0.24\columnwidth]{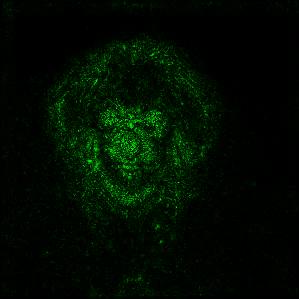}
  \label{fig:maltese_dog_blur}
  }
  \subfloat[\hspace{-0.09cm}steel~arch~bridge]{
   \includegraphics[width=0.24\columnwidth]{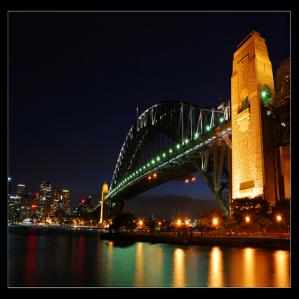}
   \label{fig:steel_arch_bridge}
   }
  \subfloat[Blur IG]{
  \includegraphics[width=0.24\columnwidth]{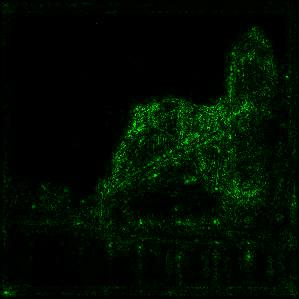}
  \label{fig:steel_arch_bridge_blur}
  }
 \captionsetup{justification=justified}
 \caption{Example images of `maltese dog' and `steel arch bridge' along with their Blur IG attribution masks.}
 \label{fig:dog_and_bridge}
\end{figure}

One application of Blur IG is it's ability to answer the question: \textbf{At what scales does the network recognize the image features relevant for the predicted class?}  Figure \ref{fig:prediction_trends} compares the relative scales at which `steel arch bridge' and `maltese dog' are classified. 
We can see from Figure~\ref{fig:steel_arch_bridge_prediction_trend} that the bulk of the Blur IG attributions occur from $\sigma=6$ to $\sigma=3$ for `steel arch bridge', while Figure~\ref{fig:maltese_dog_prediction_trend} shows that it is from $\sigma=3$ to $\sigma=0$ for `maltese dog'. Thus the dog prediction happens at far lower scales; this is probably because dog species are classified based on fine-grained features. Note that in the ImageNet data set, the objects are usually in focus, i.e., the dogs appear as large as the bridges. See figure \ref{fig:dog_and_bridge}. So we are not simply saying that dogs are smaller than bridges.

\begin{remark}[Visualizations]
For perception tasks, the attributions are almost always communicated as pictures; the scores are not directly communicated. To inspect the frequency/scale dimension, visualizing a series of saliency maps (e.g. Figure~\ref{fig:jackfruit}) is preferable to standard saliency maps (e.g. Figure~\ref{fig:comparison-of-methods}). 
The quality of the explanations do depend on the quality of visualizations~\cite{Visualizations}. %(see~\cite{Visualizations} for a longer discussion of this topic). 
\end{remark}
 
\begin{figure}
 \captionsetup[subfigure]{labelformat=empty}
 \subfloat{
   {$\qquad \qquad \qquad$ Blur IG \hspace{1.3cm}  IG (random baseline)}
%   %}   %%% 0.75in is half of figure height 1.5in
}\\\vspace{-0.7cm}\\
\subfloat{
   \raisebox{0.7in}{
   \rotatebox[origin=t]{90}{true class=`lemon'}
   }   %%% 0.75in is half of figure height 1.5in
}
 \subfloat{
   \includegraphics[width=0.45\columnwidth]{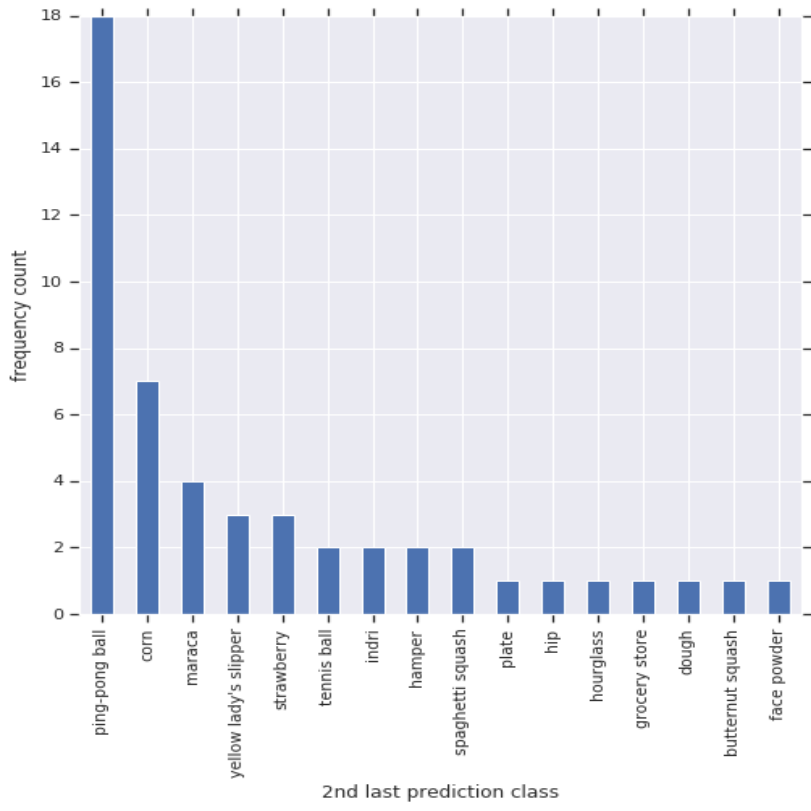}
   \label{fig:lemon_histogram_blur}
}\hspace{-1em}
 \subfloat{
   \includegraphics[width=0.45\columnwidth]{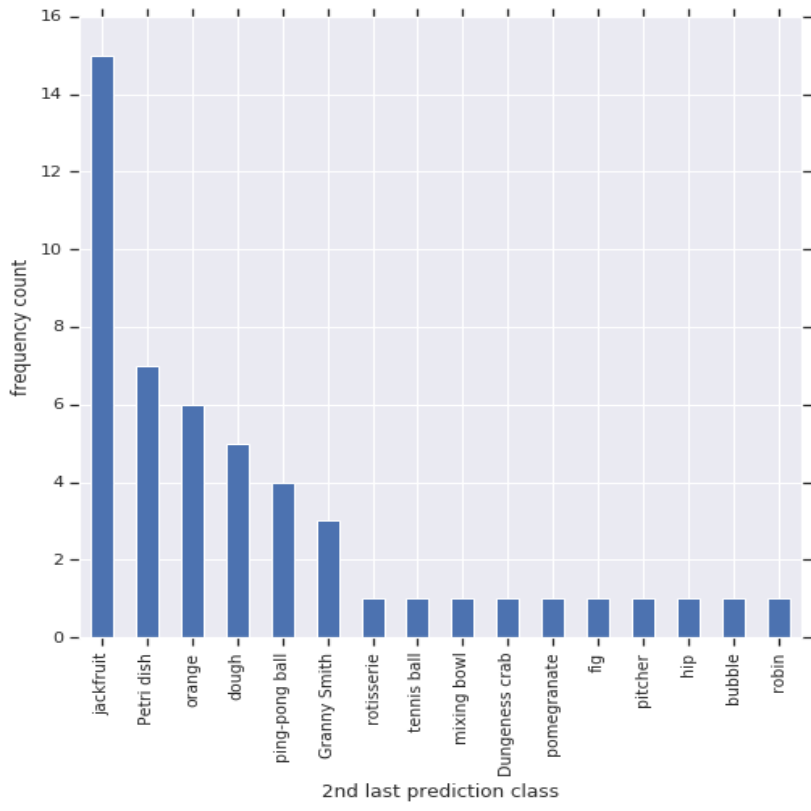}
   \label{fig:lemon_histogram_intensity}
}
\\ \vspace{-0.7cm}\\
\subfloat{
   \raisebox{0.7in}{
   \rotatebox[origin=t]{90}{true class=`junco'}
   }   %%% 0.75in is half of figure height 1.5in
}
 \subfloat{
   \includegraphics[width=0.45\columnwidth]{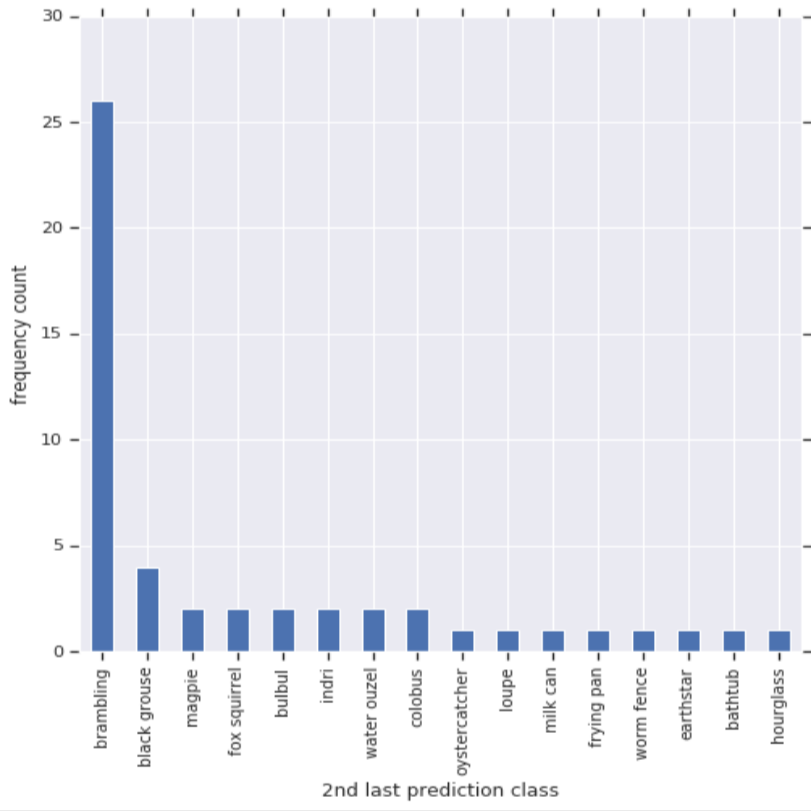}
   \label{fig:junco_histogram_blur}
}\hspace{-1em}
 \subfloat{
   \includegraphics[width=0.45\columnwidth]{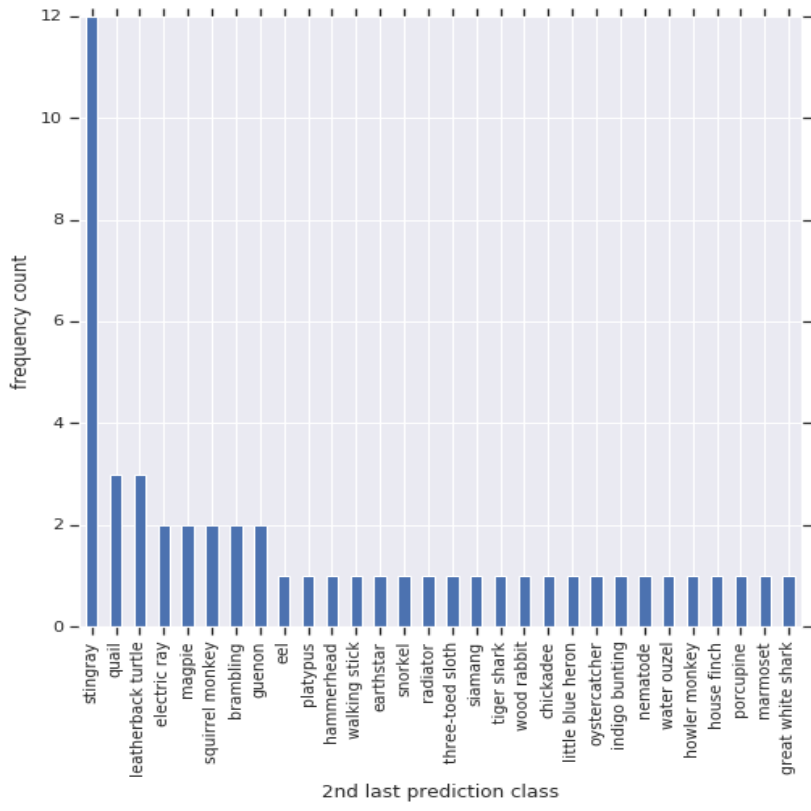}
   \label{fig:junco_histogram_intensity}
}
\\ \vspace{-0.7cm}\\
\caption{Histograms (N=50) of 2nd last predicted class along integration path of Blur IG (left) vs IG (right) for the classes `lemon' (top row) and `junco' (bottom row). %[TODO(Shawn): Normalize y axis for each class.]
}
    \label{fig:histograms}
\end{figure}

\textbf{Blur IG path produces more natural perturbations.} To test this, we sampled 100 classes, with 50 true positive images per class, and for each image found the second last label on the integration path that is not equal to the true class. If the perturbation were natural, we would expect the second last label to be something that a human would find a plausible alternate classification. See Figure \ref{fig:histograms} for examples of second last class for images from two classes, `lemon' and `junco' (a bird). The 
  most frequent 2nd last label for IG, `jackfruit' is semantically closer to `lemon' than the most frequent 2nd last class for Blur IG, `ping pong ball'. However, a ping-ping ball is  morphologically similar to a lemon. For `junco', IG's most frequent 2nd last label, `stingray', seems unrelated semantically and morphologically.
  
\textbf{Quantitative evaluation using ImageNet hierarchy.} To make the analysis objective, we use WordNet. WordNet imposes an ontology (tree) on top of the ImageNet labels. The ontology (tree) mirrors human intuition. The parent of `maltese' is `dog' and its parent is `animal' and so on. Classes that have a closer tree distance are intuitively more related. We compare the WordNet tree distance between the second last label and the true label for Blur IG, IG (random baseline), and IG (black baseline), where tree distance is defined as the sum of the distances to the lowest common ancestor of the second last label and the true label. Our analysis shows that Blur IG has the lowest average distance at 8.59, while IG with random baseline has a larger distance of 9.17, confirming that Blur IG employs more natural perturbations. To test for statistical significance, we used a paired-sample t-test with $H_0: \mu_D <= 0.25, H_1: \mu_D > 0.25$, where $\mu_D$ represents the mean difference between blur and IG random baseline label distances. With confidence level $95\%$, we reject the null hypothesis with $t = 1.674 > t_{0.05, 99} (=1.6604)$. Figure~\ref{fig:histograms} (bottom row) compares 2nd last label histograms for the `lemon' class, for which the IG path has a lower average WordNet tree distance.
 We hypothesize that the improvement in quality of the explanation masks produced by Blur IG is a result of it having a more `natural' integration path from the counterfactual input to the real input, compared with the IG integration path (see Remark~\ref{rem:baseline}, and Section~\ref{sec:discuss} for a further discussion of this). 

%\textbf{Quantitative eval using ImageNet hierarchy.} As R3 notes, Section 5.1 (lines:502-523) of the paper describes average second last label distance in the WordNet hierarchy. Blur IG with lowest average distance 8.59, considerably outperforms IG with random baseline (distance$=$9.17). %(Can we include black$+$white baseline here also?)

\textbf{Quantitative evaluation using human segmentation.} Next, we perform quantitative evaluations proposed in \cite{XRAI}. For ImageNet we use the Inception-v1 model as in \cite{XRAI}. %,cong2018review} - cited by xrai but takes up space here, we should include in final version.
Given an annotation region, the evaluation computes AUC, F1, and MAE (mean absolute error) 
of the generated saliency mask by considering pixels within the annotation to be positive and outside to be negative. In our evaluations, the annotations correspond to bounding-boxes for ImageNet, and segmentation masks for Diabetic Retinopathy (DR) (Sec.~\ref{sec:DR})%\footnote{There are 5 image-level labels (1 for No-DR, and 2$\cdots$5 indicating DR severity.) For this evaluation, we consider the No-DR class as negative, and the other 4 jointly as positive.}
, we report the average scores in  Table~\ref{tab:qnt-eval}. 
\begin{table}[h]
\small
\setlength{\tabcolsep}{2pt}
\begin{center}
\begin{tabular}{l|ccc|ccc}
\toprule
\multicolumn{1}{c|}{{Attribution}} & \multicolumn{3}{c}{{ImageNet}} & \multicolumn{3}{|c}{{Diabetic Retinopathy}} \\
\multicolumn{1}{c|}{{method}} & AUC $\uparrow$ & F1 $\uparrow$ & MAE $\downarrow$ & AUC $\uparrow$ & F1 $\uparrow$ & MAE $\downarrow$ \\
\midrule
XRAI & 0.836 & 0.786 & 0.149 &
%XRAI & \textbf{0.836} & \textbf{0.786} & \textbf{0.149} &
        0.805 & 0.285 & 0.068 \\
GradCAM & 0.742 & 0.715 & 0.194 &
         0.817 & 0.249 & 0.058 \\
IG (random-4) & 0.709 & 0.674 & 0.223 &
            0.827 & 0.344 & 0.060 \\
IG (black) & 0.710 & 0.674 & 0.219 &
         0.828 & 0.307 & 0.062 \\
IG (black$+$white) & 0.729 & 0.681 & 0.216 &
         0.818 & 0.296 & 0.062 \\
\textbf{Blur IG (ours)} & 0.738 & 0.693 & 0.209 &
         0.831 & 0.293 & 0.061 \\
%         \textbf{0.831} & 0.293 & 0.061` \\
% \midrule
\bottomrule
\end{tabular}
\caption[qnt-eval]{
Average F1, AUC, and MAE scores for different explanation methods on images from ImageNet validation set (N=9684), and Diabetic Retinopathy dataset (N=141). ($\uparrow$ indicates higher is better, $\downarrow$ indicates lower is better) %, \textbf{bold} is best value).
}\label{tab:qnt-eval}
\end{center}
\vspace{-0.2cm}
\end{table}

%Results from Table~\ref{tab:qnt-eval} indicate that on ImageNet images BlurIG outperforms IG with different baselines, and XRAI which is segmentation based has explanations closely matching human interpretability. However, when we look at Diabetic Retinopathy, BlurIG followed by IG perform better as they are capable of also identifying finer details. (Mukund: Do we want to bold numbers? On DR IG is better on F1 and gradCAM on MAE, so not highlighting.)
GradCAM and XRAI outperform BlurIG on Imagenet, but BlurIG does better on the DR task. There's a difference between the methods on the two tasks since XRAI (and GradCAM) have lower resolution explanations. XRAI 
employs image segmentation (that is oblivious to the model), and then groups the IG (Black+White) attributions by these segments. This trades faithfulness to the model’s behavior for better visual coherence on natural images (ImageNet), but makes the results worse for tasks like DR that do not involve “natural” features. Finally, we note that the segmentation idea is orthogonal to the attribution technique, i.e., XRAI could use BlurIG instead of IG (Black+White).

\textbf{Note on biases.}
%\vspace{-0.1cm}
Different methods have different biases. We note that IG has a bias towards color depending on the color contrast (or the lack thereof) between the class of interest and the baseline whereas BlurIG has a bias towards shape. This is illustrated in Fig.~\ref{fig:bias} below.
\vspace{-0.4cm}
\begin{figure}[h]
    \centering
    \subfloat[Original]{\includegraphics[width=0.124\textwidth]{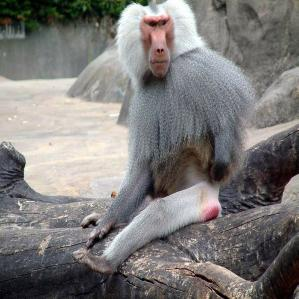}}
    \subfloat[IG (grayscale)]{\includegraphics[width=0.124\textwidth]{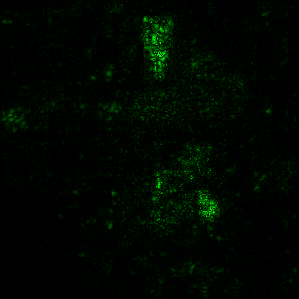}}
    %\vspace{-1\baselineskip}
    \subfloat[IG (black)]{\includegraphics[width=0.124\textwidth]{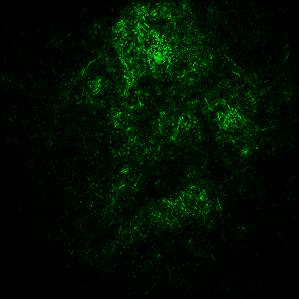}}
    \subfloat[Blur IG]{\includegraphics[width=0.124\textwidth]{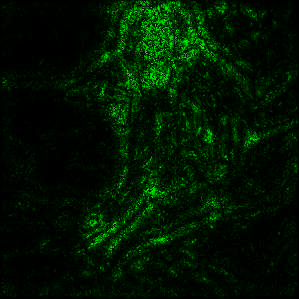}}
\caption{\small{[Bias] Examples to show that IG has bias towards color and Blur IG has a bias towards shape. 
The 4 images (left to right) correspond to the original, followed by saliency masks for IG with a grayscale version of the input image as baseline,  
IG with a black baseline, and Blur IG. Notice that IG (grayscale) 
focuses on
the baboon's pink face and bottom, and IG (black) emphasizes light features. In contrast, Blur IG emphasizes its overall shape, paying close attention to the shape of the head and limbs. 
}}
\label{fig:bias}
\end{figure}

%----------------------------------

 \subsection{Diabetic Retinopathy}
 \label{sec:DR}
 
\begin{figure}%[ht!]
 \captionsetup[subfigure]{labelformat=empty}
 \subfloat{
   {$\quad$ Original \hspace{0.5cm} GradCAM \hspace{0.8cm} IG \hspace{1.0cm} Blur IG}
%   %}   %%% 0.75in is half of figure height 1.5in
}\\\vspace{-0.6cm}\\
 \subfloat{
   \includegraphics[width=0.24\columnwidth]{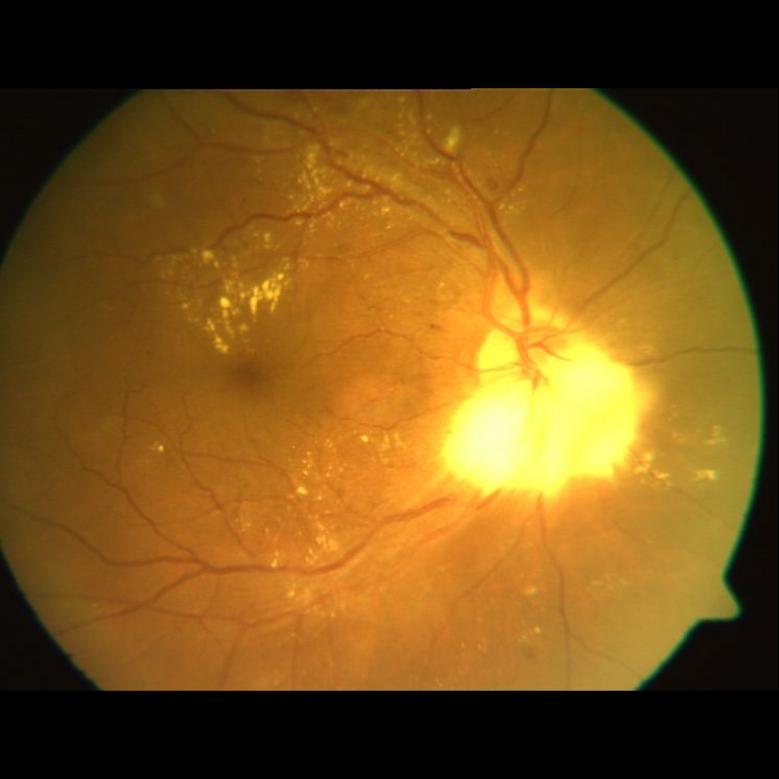}
}\hspace{-0.25cm}
 \subfloat{
   \includegraphics[width=0.24\columnwidth]{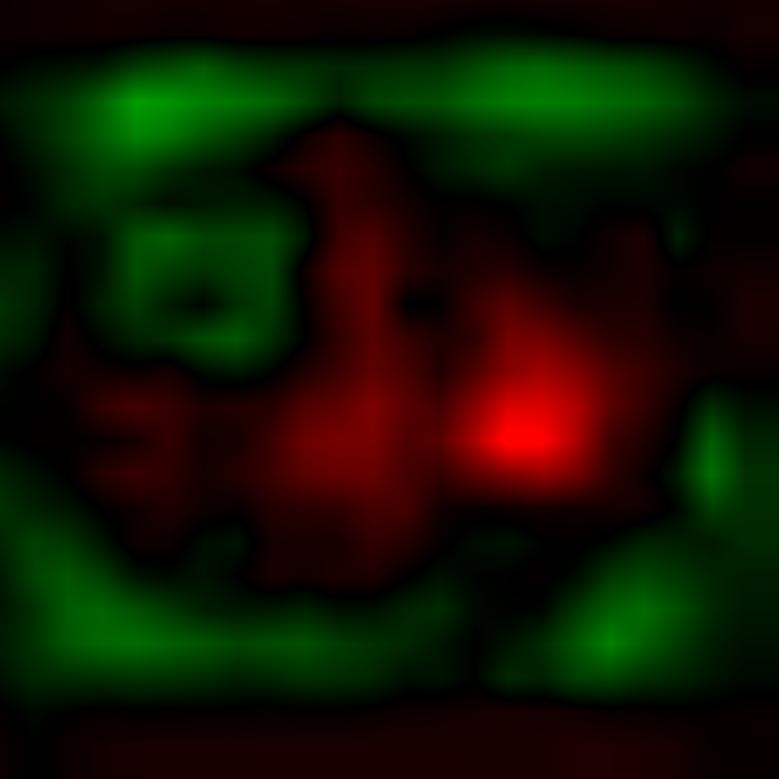}
}\hspace{-0.24cm}
 \subfloat{
   \includegraphics[width=0.24\columnwidth]{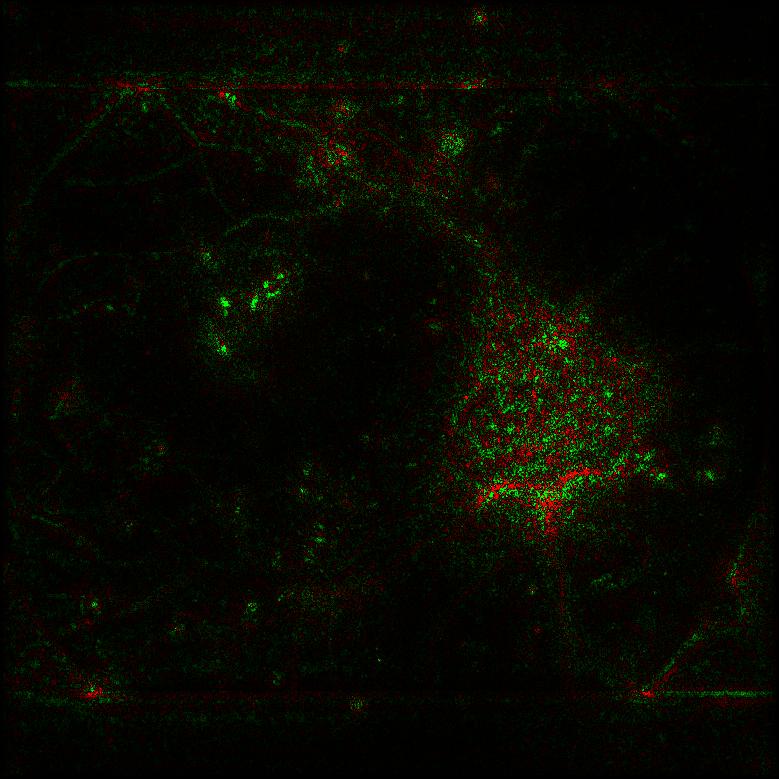}
}\hspace{-0.25cm}
 \subfloat{
   \includegraphics[width=0.24\columnwidth]{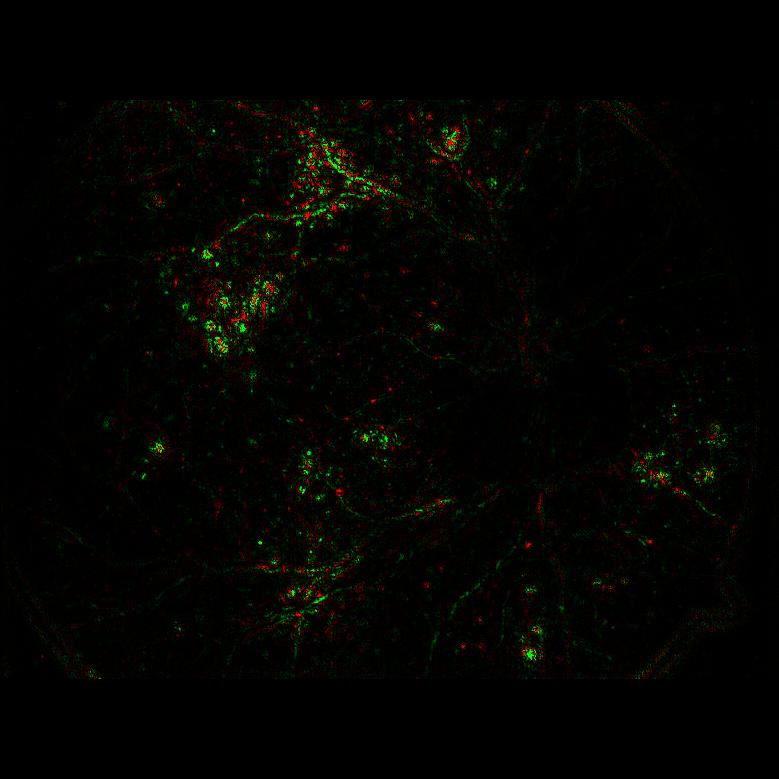}
}
\\\vspace{-0.75cm}\\
\subfloat{
   \includegraphics[width=0.24\columnwidth]{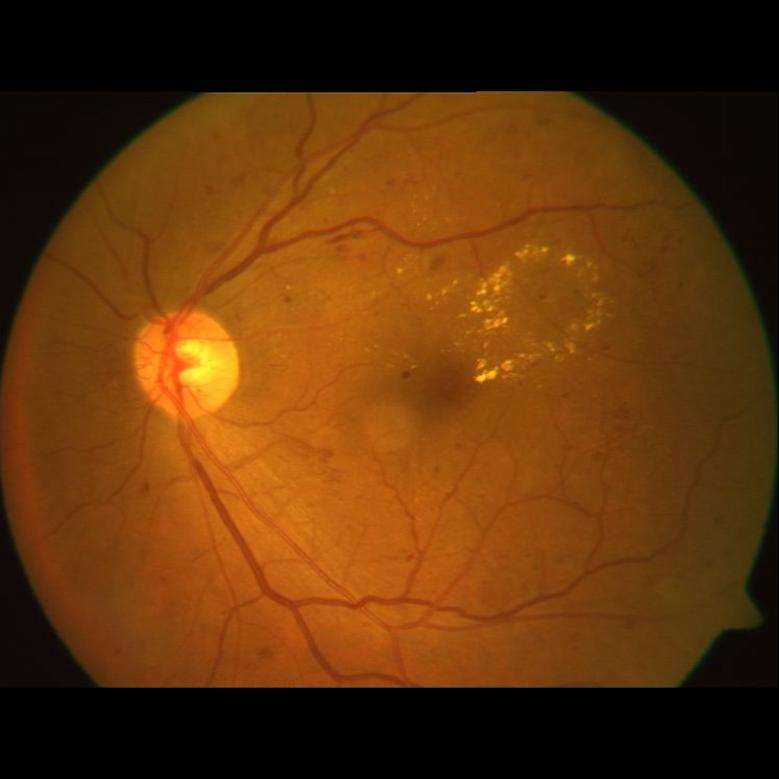}
}\hspace{-0.25cm}
 \subfloat{
   \includegraphics[width=0.24\columnwidth]{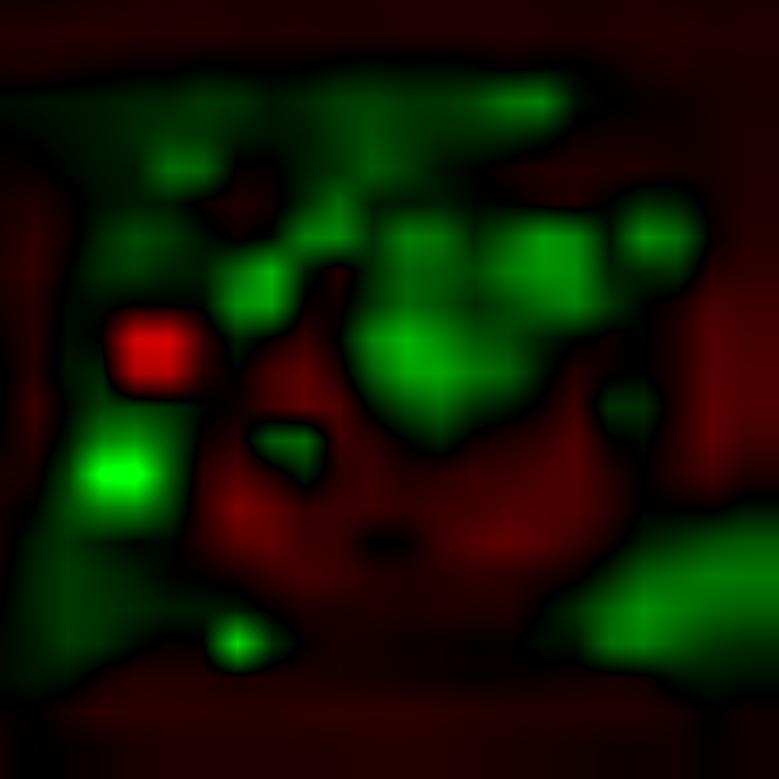}
}\hspace{-0.24cm}
 \subfloat{
   \includegraphics[width=0.24\columnwidth]{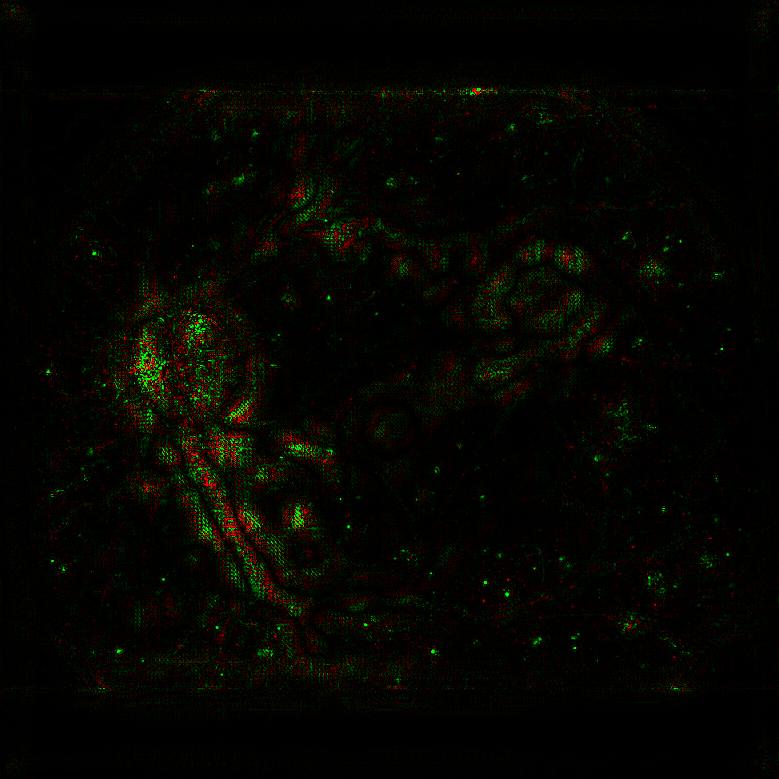}
}\hspace{-0.25cm}
 \subfloat{
   \includegraphics[width=0.24\columnwidth]{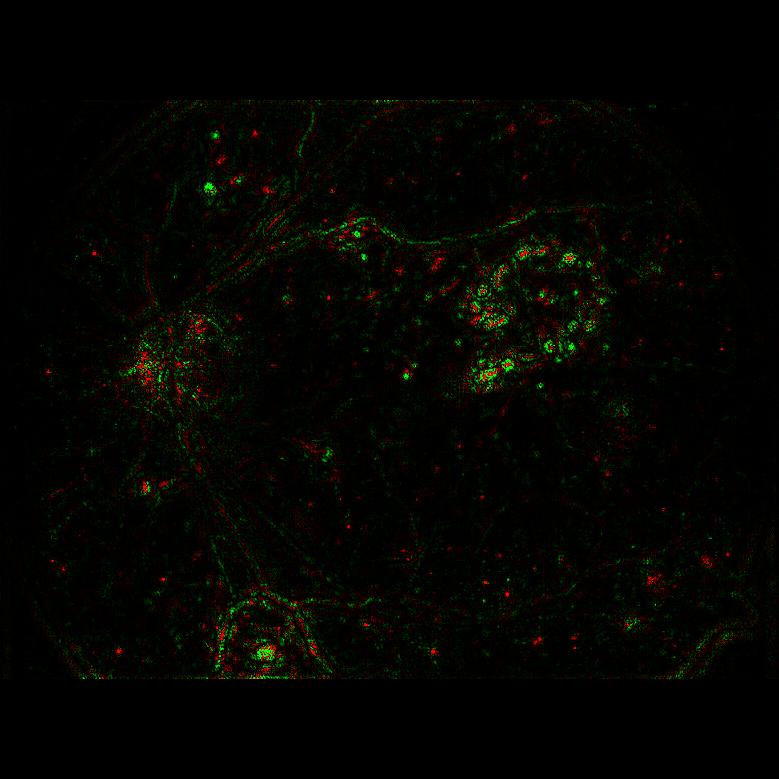}
}
\\\vspace{-0.3cm}
\caption{Comparison of GradCAM, random baseline IG, and Blur IG on retina images diagnosed with diabetic retinopathy. Green and red colors indicate positive and negative attributions respectively.}
\label{fig:dr}
\end{figure}

We also apply Blur IG to a Diabetic Retinopathy prediction task. %We apply Blur IG to an object recognition task. 
Specifically, we study the model from~\cite{jama-dr} that uses the Inception-V4 architecture~\cite{Inceptionv4}. Figure~\ref{fig:dr} compares GradCAM, random baseline IG, and Blur IG on retina images diagnosed with diabetic retinopathy (DR). The first image contains hard exudates (small white or yellowish deposits), retinal hemorrhages (dark spots), and micro-aneurysms (small, round, dark spots), with a diagnosis of moderate DR determined by retina specialists. Notice that the GradCAM explanation is too coarse-grained to be appropriate for explanation of DR classification. The random baseline IG explanation correctly attributes the hard exudates at the left and top of the image, but misattributes the bright spot on the optic disk, which is a result of a camera artifact (brightness oversaturation). In contrast, the Blur IG explanation correctly attributes the DR lesions and ignores the bright spot.
The second image contains an example retina image diagnosed with proliferative DR. Critical to this diagnosis is the presence of neovascularization of the optic disc (NVD). GradCAM fails to attribute the NVD lesion (its attribution in that region is purely negative, which indicates that the lesion is not relevant to the classification of DR). Both random baseline IG and Blur IG attribute the NVD, but random baseline IG also seems to highlight numerous spurious spots that does not seem to have clinically relevant features. In addition, Blur IG attributes blood vessels that possibly contain venous beading, another lesion associated with proliferative DR.

%----------------------------------

\subsection{Audio Classification}
 \label{sec:audio}
 
We apply Blur IG to study a CNN model~\cite{AudioCNN} trained on AudioSet~\cite{AudioSet} audio event recognition, i.e., the task of predicting types of audio events from an ontology of
 %TODO(subha): check if it is 527
635 classes that range across human sounds, animal sounds, musical sounds, sounds made by objects, etc. The CNN is a modified  ResNet-50~\cite{he2016deep}. The model takes the spectrogram of the waveform. (A spectrogram is a visual representation of the spectrum of frequencies of a signal as it varies with time.) The audio is processed as non-overlapping 960ms frames. These are decomposed with a short-time Fourier transform applying 25 ms windows every 10 ms. The resulting spectrogram is integrated into
64 mel-spaced frequency bins, and the magnitude of each bin is log transformed. This gives log-mel spectrogram patches of 96X64 bins that form
the input. The output of the model is a multilabel classification. For instance, an audio sample of a violin is expected to be classified as a violin but also as a bow-stringed instrument, and as music. We apply this model on audio samples publicly available from the Freesound audio tagging challenge~\cite{fonseca2018general}.

% \begin{figure}%
%    \centering
%    \includegraphics[width=0.45\textwidth]{figures/audio/blur_on_audio2.png}
%    \caption{\small{%\todo{SV: Placeholder}
%    Visualizing the evolution of class prediction probabilities along the blur integration path from a violin audio sample. Y axis shows the top 16 predicted labels sorted by the mean confidence for the class across the entire blur path. Color indicates the actual confidence score, darker is higher confidence. Initially model has higher confidence on violin and string instrument classes. With increased blur, confidence shifts towards synthesizer, keyboard, french horn, and then singing bowl.}}%
%    \label{fig:violin_prediction_trend}%
%\end{figure}

 \begin{figure}[!b]%
 \vspace{-0.4cm}
    \centering
    \includegraphics[width=0.45\textwidth]{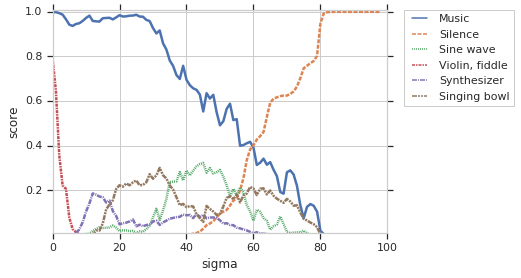}%
    \caption{\small{%\todo{SV: Placeholder}
    Visualizing the evolution of class prediction probabilities for prominent categories along the blur integration path from a violin audio sample. Y axis shows the confidence score, and X axis the sigma for the gaussian blur kernel. Color indicates the class. Initially model has higher confidence on violin (and music) class. With increased blur, confidence shifts towards synthesizer, singing bowl, sine wave, and then silence.}}%
    \label{fig:violin_prediction_trend_lines}%
\end{figure}

 \begin{figure}%
    \centering
    \includegraphics[width=0.45\textwidth]{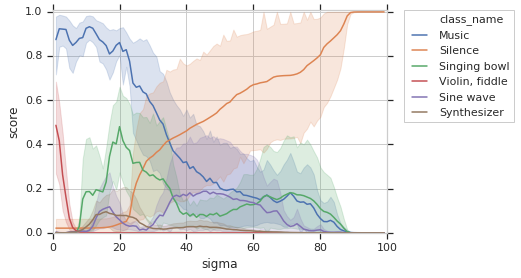}%
    \caption{\small{%\todo{SV: Placeholder}
    Visualizing the evolution of class prediction probabilities for prominent categories along the blur integration path from a few (7) violin audio samples. Y axis shows the confidence score, and X axis the sigma for the gaussian blur kernel. Color indicates the class. Initially model has higher confidence on violin and string instrument classes. With increased blur, confidence shifts towards singing bowl, sine wave, and then silence.}}%
    \label{fig:agg_violin_prediction_trend_lines}%
\end{figure}

%\paragraph{Evolution of prediction.}
\textbf{Evolution of prediction.}
Figure~\ref{fig:violin_prediction_trend_lines} shows the evolution of predictions along the blur path on a violin audio sample. As the blur increases, prediction transforms from violin to synthesizer and then to singing bowl. The synthesizer and the singing bowl are more smooth sounding than the violin, indicating that the path is `natural', increasing the likelihood of a good explanation. 

\begin{figure}%
    \centering
    \includegraphics[width=0.45\textwidth]{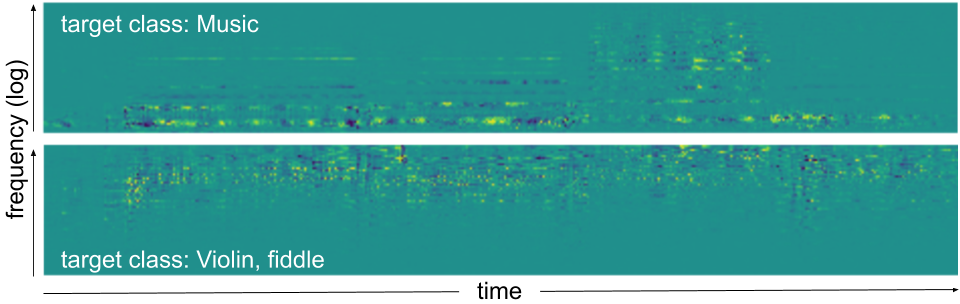}%
    \caption{\small{
    %\todo{SV: Placeholder}
    Blur IG explanations on a violin audio sample with target class as `Music' (\textit{top}) and target class as `Violin, fiddle' (\textit{below}). Yellow indicates positive gradients and blue the negative gradients. Explanation for class Music focuses on the lower frequencies while explanation for `Violin, fiddle' is on the higher frequencies.% (TODO(subha): Use same color scheme as imagenet)
    }}%
    \label{fig:violin_vs_music}%
\end{figure}

\begin{figure}%
    \subfloat[Target class: `Music']{\includegraphics[width=0.5\columnwidth,height=3.1cm]{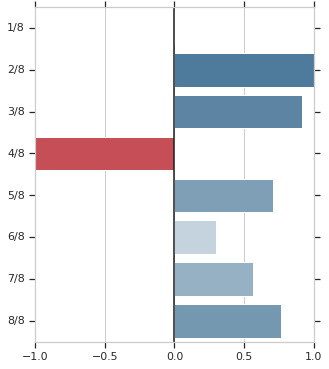}}
    \subfloat[Target class: `Violin, fiddle`]
    {\includegraphics[width=0.5\columnwidth,height=3.1cm]{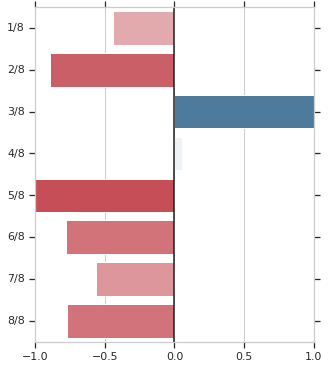}}
%    \centering
%    \includegraphics[width=0.21\textwidth]{figures/audio/music_class_agg7_samples.png}% \\
%    \includegraphics[width=0.21\textwidth]{figures/audio/violin_class_agg7_samples.png}%
    \caption{\small{
    %\todo{SV: Placeholder}
    Aggregation of Blur IG contributions on 7 violin audio samples with target class as `Music' (\textit{left}) and target class as `Violin, fiddle' (\textit{right}). Y axis depicts the frequency bins, and X axis the integrated gradients. Blue indicates positive gradients/contributions and red the negative gradients. Explanation for class Music shows positive contributions from the lower frequencies while explanation for `Violin, fiddle' shows positive contributions only from the higher frequencies.}}%
    \label{fig:violin_vs_music_agg}%
\end{figure}

%\paragraph{Class conditioning.} 
\textbf{Class conditioning.} Next, we show how Blur IG can be used to identify insights. We examine Blur IG explanations for the same audio sample. The model is a multi-label classifier and tags the piece both as `Music' and as `Violin'. We use Blur IG to explain both predictions. 
We would expect the model to predict music \emph{because} it predicts violin, and therefore, the attributions for the classes to match. However, 
we find that the model looks at \emph{different} frequencies for the two classes! Figure~\ref{fig:violin_vs_music} shows the attributions. The model looks at lower frequencies for music than violin.  For target class violin, the sum of the Blur IG explanations for the higher frequencies (top half across entire clip) is positive while that of the lower frequencies is negative (ratio is $\approx 19:-1$). Whereas for class music, the sum of contributions of the higher frequencies is negative while that of the lower frequencies is positive (ratio is $\approx -33:1$). This is also easily observed in Figure~\ref{fig:violin_vs_music_agg} which aggregates attributions within different frequency bins (for multiple violin samples). This insight demonstrates the utility of inspecting the frequency domain.  

%----------------------------------

%\vspace{-2pt}
\section{Discussion}
 \label{sec:discuss}
%\vspace{-2pt}
Explanations are associated with perturbations. For instance, the explanation that `Event A caused Event B' implies that \emph{had Event A not occurred}, Event B would not either; the phrase in italics is the perturbation associated with the explanation. The Counterfactual Theory of Causation by \cite{Lewis} formalizes this argument; here, the word counterfactual is synonymous with perturbation.
 
 Feature importance/attributions are a form of explanation and indeed all the techniques that compute attributions use some type of perturbation. A good attribution technique should have two characteristics:
\begin{itemize}[noitemsep,topsep=0pt]
 \item For the attribution to produce a human intelligible explanation, we like the perturbation to be one that involves changing a human intelligible feature. 
 \item To ensure that the explanation does not produce artifacts, we would like the perturbations that only destroy information. 
\end{itemize}
 

 Blur Integrated Gradients is a \emph{best-effort} to satisfy both conditions. Blur Integrated Gradient relies on Gaussians and Laplacians of the Gaussian to construct explanations. It is well-known that operators built using Gaussians and Laplacians of Gaussians detect edges, blogs and textures (\eg\cite{Lindeberg}), \ie, features that are human intelligible. Indeed, there is also biological evidence that humans rely on Gaussians and Laplacians of Gaussians to perform visual and auditory tasks~\cite{Vision1,Hearing,zhang2018unreasonable}. 
 
 On the computer vision front, Gaussian and Laplacians of Gaussians are workhorses of multi-scale processing. Convolutional deep neural networks used for perception tasks are built with this intuition. As the Inception paper~\cite{Inception} says: \emph{`To optimize quality, the architectural
decisions were based on ... and the intuition of multi-scale processing.'} By choosing an explanation method that fits the \emph{form} of deep networks and what humans consider features, we hope to generate explanations that are simultaneously faithful to the network and intelligible. 
As a side-effect, we are also able to generate explanations in scale/frequency. And we show with examples in vision and audio.

%----------------------------------

{\small
\bibliographystyle{ieee_fullname}
\bibliography{main}

\begin{thebibliography}{10}\itemsep=-1pt

\bibitem{AnkurGit}
github.com/ankurtaly/integrated-gradients.

\bibitem{AS74}
R.~J. Aumann and L.~S. Shapley.
\newblock {\em Values of Non-Atomic Games}.
\newblock Princeton University Press, 1974.

\bibitem{BSHKHM10}
David Baehrens, Timon Schroeter, Stefan Harmeling, Motoaki Kawanabe, Katja
  Hansen, and Klaus-Robert M{\"u}ller.
\newblock How to explain individual classification decisions.
\newblock {\em Journal of Machine Learning Research}, pages 1803--1831, 2010.

\bibitem{BMBMS16}
Alexander Binder, Gr{\'e}goire Montavon, Sebastian Bach, Klaus-Robert
  M{\"u}ller, and Wojciech Samek.
\newblock Layer-wise relevance propagation for neural networks with local
  renormalization layers.
\newblock {\em CoRR}, 2016.

\bibitem{fong}
R. Fong, M. Patrick, and A. Vedaldi.
\newblock Understanding deep networks via extremal perturbations and smooth
  masks.
\newblock In {\em Proceedings of the International Conference on Computer
  Vision ({ICCV})}, 2019.

\bibitem{fonseca2018general}
Eduardo Fonseca, Manoj Plakal, Frederic Font, Daniel~PW Ellis, Xavier Favory,
  Jordi Pons, and Xavier Serra.
\newblock General-purpose tagging of freesound audio with audioset labels: Task
  description, dataset, and baseline.
\newblock {\em arXiv preprint arXiv:1807.09902}, 2018.

\bibitem{Friedman}
Eric~J. Friedman.
\newblock Paths and consistency in additive cost sharing.
\newblock {\em International Journal of Game Theory}, 32(4):501--518, 2004.

\bibitem{AudioSet}
Jort~F. Gemmeke, Daniel P.~W. Ellis, Dylan Freedman, Aren Jansen, Wade
  Lawrence, R.~Channing Moore, Manoj Plakal, and Marvin Ritter.
\newblock Audio set: An ontology and human-labeled dataset for audio events.
\newblock In {\em Proc. IEEE ICASSP 2017}, New Orleans, LA, 2017.

\bibitem{he2016deep}
Kaiming He, Xiangyu Zhang, Shaoqing Ren, and Jian Sun.
\newblock Deep residual learning for image recognition.
\newblock In {\em Proceedings of the IEEE conference on computer vision and
  pattern recognition}, pages 770--778, 2016.

\bibitem{AudioCNN}
Shawn Hershey, Sourish Chaudhuri, Daniel P.~W. Ellis, Jort~F. Gemmeke, Aren
  Jansen, Channing Moore, Manoj Plakal, Devin Platt, Rif~A. Saurous, Bryan
  Seybold, Malcolm Slaney, Ron Weiss, and Kevin Wilson.
\newblock Cnn architectures for large-scale audio classification.
\newblock In {\em International Conference on Acoustics, Speech and Signal
  Processing (ICASSP)}. 2017.

\bibitem{XRAI}
Andrei Kapishnikov, Tolga Bolukbasi, Fernanda Viegas, and Michael Terry.
\newblock Xrai: Better attributions through regions.
\newblock In {\em The IEEE International Conference on Computer Vision (ICCV)},
  October 2019.

\bibitem{Koenderink}
Jan~J. Koenderink.
\newblock The structure of images.
\newblock {\em Biological Cybernetics}, 50(5):363--370, Aug 1984.

\bibitem{Lewis}
David~K. Lewis.
\newblock {\em Counterfactuals}.
\newblock Blackwell, 1973.

\bibitem{Lindeberg}
T. Lindeberg.
\newblock Scale-space for discrete signals.
\newblock {\em IEEE Trans. Pattern Anal. Mach. Intell.}, 12(3):234--254, Mar.
  1990.

\bibitem{Blob}
Tony Lindeberg.
\newblock Detecting salient blob-like image structures and their scales with a
  scale-space primal sketch: A method for focus-of-attention.
\newblock {\em International Journal of Computer Vision}, 11(3):283--318, Dec
  1993.

\bibitem{scale-space}
Tony Lindeberg.
\newblock {\em Scale-Space Theory in Computer Vision}.
\newblock Kluwer Academic Publishers, Norwell, MA, USA, 1994.

\bibitem{Vision1}
Tony Lindeberg.
\newblock A computational theory of visual receptive fields.
\newblock {\em Biological Cybernetics}, 107(6):589--635, Dec 2013.

\bibitem{Hearing}
Tony Lindeberg and Anders Friberg.
\newblock Scale-space theory for auditory signals.
\newblock In Jean-Fran{\c{c}}ois Aujol, Mila Nikolova, and Nicolas Papadakis,
  editors, {\em Scale Space and Variational Methods in Computer Vision}, pages
  3--15, Cham, 2015. Springer International Publishing.

\bibitem{SIFT}
David~G. Lowe.
\newblock Distinctive image features from scale-invariant keypoints.
\newblock {\em Int. J. Comput. Vision}, 60(2):91--110, Nov. 2004.

\bibitem{Lundberg2017AUA}
Scott Lundberg and Su-In Lee.
\newblock A unified approach to interpreting model predictions.
\newblock In {\em NIPS}, 2017.

\bibitem{LL17}
Scott~M Lundberg and Su-In Lee.
\newblock A unified approach to interpreting model predictions.
\newblock In I. Guyon, U.~V. Luxburg, S. Bengio, H. Wallach, R. Fergus, S.
  Vishwanathan, and R. Garnett, editors, {\em Advances in Neural Information
  Processing Systems 30}, pages 4768--4777. Curran Associates, Inc., 2017.

\bibitem{deepdream}
Alexander Mordvintse, Christopher Olah, and Mike Tyka.
\newblock {Inceptionism: Going Deeper into Neural Networks}, 2015.

\bibitem{DBLP:conf/icml/2017}
Doina Precup and Yee~Whye Teh, editors.
\newblock {\em Proceedings of the 34th International Conference on Machine
  Learning, {ICML} 2017, Sydney, NSW, Australia, 6-11 August 2017}, volume~70
  of {\em Proceedings of Machine Learning Research}. {PMLR}, 2017.

\bibitem{Lime}
Marco~T{\'{u}}lio Ribeiro, Sameer Singh, and Carlos Guestrin.
\newblock "why should {I} trust you?": Explaining the predictions of any
  classifier.
\newblock {\em CoRR}, abs/1602.04938, 2016.

\bibitem{ILSVRC15}
Olga Russakovsky, Jia Deng, Hao Su, Jonathan Krause, Sanjeev Satheesh, Sean Ma,
  Zhiheng Huang, Andrej Karpathy, Aditya Khosla, Michael Bernstein,
  Alexander~C. Berg, and Li Fei-Fei.
\newblock {ImageNet Large Scale Visual Recognition Challenge}.
\newblock {\em International Journal of Computer Vision (IJCV)}, pages
  211--252, 2015.

\bibitem{Selvaraju}
Ramprasaath~R. Selvaraju, Michael Cogswell, Abhishek Das, Ramakrishna Vedantam,
  Devi Parikh, and Dhruv Batra.
\newblock Grad-cam: Visual explanations from deep networks via gradient-based
  localization.
\newblock In {\em The IEEE International Conference on Computer Vision (ICCV)},
  Oct 2017.

\bibitem{SGSK16}
Avanti Shrikumar, Peyton Greenside, and Anshul Kundaje.
\newblock Learning important features through propagating activation
  differences.
\newblock In Precup and Teh \cite{DBLP:conf/icml/2017}, pages 3145--3153.

\bibitem{SVZ13}
Karen Simonyan, Andrea Vedaldi, and Andrew Zisserman.
\newblock Deep inside convolutional networks: Visualising image classification
  models and saliency maps.
\newblock {\em CoRR}, 2013.

\bibitem{GuidedBackProp}
J.T. Springenberg, A. Dosovitskiy, T. Brox, and M. Riedmiller.
\newblock Striving for simplicity: The all convolutional net.
\newblock In {\em ICLR (workshop track)}, 2015.

\bibitem{SDBR14}
Jost~Tobias Springenberg, Alexey Dosovitskiy, Thomas Brox, and Martin~A.
  Riedmiller.
\newblock Striving for simplicity: The all convolutional net.
\newblock {\em CoRR}, 2014.

\bibitem{STY17}
Mukund Sundararajan, Ankur Taly, and Qiqi Yan.
\newblock Axiomatic attribution for deep networks.
\newblock In Precup and Teh \cite{DBLP:conf/icml/2017}, pages 3319--3328.

\bibitem{Visualizations}
Mukund Sundararajan, Jinhua Xu, Ankur Taly, Rory Sayres, and Amir Najmi.
\newblock Exploring principled visualizations for deep network attributions.
\newblock 2019.

\bibitem{Inceptionv4}
Christian Szegedy, Sergey Ioffe, Vincent Vanhoucke, and Alexander~A. Alemi.
\newblock Inception-v4, inception-resnet and the impact of residual connections
  on learning.
\newblock In {\em Proceedings of the Thirty-First AAAI Conference on Artificial
  Intelligence}, AAAI'17, pages 4278--4284. AAAI Press, 2017.

\bibitem{Inception}
Christian Szegedy, Wei Liu, Yangqing Jia, Pierre Sermanet, Scott~E. Reed,
  Dragomir Anguelov, Dumitru Erhan, Vincent Vanhoucke, and Andrew Rabinovich.
\newblock Going deeper with convolutions.
\newblock {\em CoRR}, 2014.

\bibitem{Inception-v2}
Christian Szegedy, Vincent Vanhoucke, Sergey Ioffe, Jonathon Shlens, and
  Zbigniew Wojna.
\newblock Rethinking the inception architecture for computer vision.
\newblock In {\em Proceedings of IEEE Conference on Computer Vision and Pattern
  Recognition,}, 2016.

\bibitem{jama-dr}
Gulshan V, Peng L, Coram M, and et al.
\newblock Development and validation of a deep learning algorithm for detection
  of diabetic retinopathy in retinal fundus photographs.
\newblock {\em JAMA}, 316(22):2402--2410, 2016.

\bibitem{Witkin}
Andrew~P. Witkin.
\newblock Scale-space filtering.
\newblock In {\em Proceedings of the Eighth International Joint Conference on
  Artificial Intelligence - Volume 2}, IJCAI'83, pages 1019--1022, San
  Francisco, CA, USA, 1983. Morgan Kaufmann Publishers Inc.

\bibitem{zhang2018unreasonable}
Richard Zhang, Phillip Isola, Alexei~A Efros, Eli Shechtman, and Oliver Wang.
\newblock The unreasonable effectiveness of deep features as a perceptual
  metric.
\newblock In {\em Proceedings of the IEEE Conference on Computer Vision and
  Pattern Recognition}, pages 586--595, 2018.

\end{thebibliography}
}
\newpage

%\section*{Appendix}
%
% \begin{figure}[h]%
%    \centering
%    \includegraphics[width=0.45\textwidth]{figures/audio/agg_predtrend_violin_appendix_moreclasses.png}%
%    \caption{\small{%\todo{SV: Placeholder}
%    Visualizing the evolution of class prediction probabilities for prominent categories along the blur integration path from a few (7) violin audio samples. Y axis shows the confidence score, and X axis the sigma for the Gaussian blur kernel. Color indicates the class. Initially model has higher confidence on violin and musical instrument classes. With increased blur, confidence shifts towards singing bowl, sine wave, and then silence.}}%
%    \label{fig:agg_violin_prediction_trend_lines_app}%
%\end{figure}

\end{document}